\documentclass[letterpaper]{article} %% AAAI 2013
\usepackage{myijcai19}
\usepackage{times}
\usepackage{courier}
\usepackage{theorem}
\usepackage{xspace}
\usepackage{boxedminipage}
\usepackage{graphicx}
\usepackage{amssymb}
\usepackage{amsmath}
\usepackage{amsfonts}
\usepackage{alltt}
\usepackage{array}
\usepackage{latexsym}
\usepackage{color}
\usepackage{url}
\usepackage{soul}
\usepackage{tabularx}
\usepackage{amssymb}
\newif\ifcomments
\commentsfalse

\ifcomments
\newcommand{\commentsm}[1]{\textcolor{blue}{(SM: #1)}}
\newcommand{\commentms}[1]{\textcolor{magenta}{(MS: #1)}}
\newcommand{\old}[1]{\textcolor{red}{\sout{#1}}}

\else
 \newcommand{\commentsm}[1]{}
 \newcommand{\commentms}[1]{}
 \newcommand{\old}[1]{}
 
\fi

\newcommand{\fs}[1]{\fontsize{#1}{#1}\selectfont}

\makeatletter
\newcommand{\nbcite}{\def\citeauthoryear##1##2{\def\@thisauthor{##1}%
\ifx \@lastauthor \@thisauthor \relax \else##1 \fi ##2}\@nbcite}
\def\citeS{\@ifnextchar[{\@jbciteS}{\@jbciteS[]}}
\def\@jbciteS[#1]#2{%
\ifthenelse{\equal{#1}{}}{%
\citeauthor{#2}'s (\citeyear{#2})}{%
\citeauthor{#2}'s #1 (\citeyear{#2})}}
\makeatother

\newcommand{\eat}[1]{}

\def\noop{\emph{discard}\xspace}

\def\next{\textbf{next}} 
 
\def\shop2{\textbf{{\fs{8.5}\textsc{SHOP2}}}}

\def\next{\textbf{next}}

\def\iff{\mbox{iff}}

\newcounter{annocount}
  {\end{alltt}}

\theoremstyle{plain} 
\newtheorem{definition}{Definition}

\begin{document}

%\title{(The) Dimensions of Empathetic Planning}
%\title{Empathetic Planning and Plan Recognition for Human Well-being}
\title{Towards Empathetic Planning\footnote{This work was presented at the 2nd Workshop on Humanizing AI (HAI) at IJCAI'19 in Macao, China.}}

%\\ and its Role in Building Assistive Agents}

%\title{Multi-Agent Plan Recognition with Temporal Actions as Planning}

%\author{Submission \#5636}

%\author{Maayan \And Sheila}

\author{
Maayan Shvo
\And
Sheila A. McIlraith
\affiliations
Department of Computer Science, University of Toronto, Canada\\
\emails
\{maayanshvo, sheila\}@cs.toronto.edu
}

% \author{Maayan Shvo  \\ Department of Computer Science, University of Toronto \\ Toronto, Canada \\
%  \and
%  Sheila A. McIlraith \\  Department of Computer Science, University of Toronto \\ Toronto, Canada}

% TO REMOVE THE COVER PAGE JUST COMMENT OUT THE FOLLOWING:
%\input{coverpage.tex} 

%\nocopyright
\maketitle

% Squeezing
%\setlength\floatsep{0.2cm}
%% ADD This later if worried about Space
\setlength\textfloatsep{0.4cm}
\setlength\theorempreskipamount{6pt plus 1pt minus 0.5pt}

\newcommand{\Sys}{\ensuremath{\Sigma}\xspace}
\def\next{\raisebox{1pt}{\begin{footnotesize}\ensuremath{\bigcirc}\end{footnotesize}}}
\newcommand{\eventually}{\begin{large}\ensuremath{\lozenge}\end{large}}
\newcommand{\prefeq}{\ensuremath{\preceq}\xspace}
\newcommand{\pref}{\ensuremath{\prec}\xspace}

\def\noops{\emph{discard}}
\def\noop{\emph{discard}\xspace}

\setcounter{secnumdepth}{2}

\begin{abstract}

\eat{Every compassionate and functioning society requires its members to have a capacity to adopt others' perspectives. }

%
%
% / to adtop others' perspectives.
%
% The ability to empathize is critical critical human 
% Empathy is a critical aspect of human communication and every compassionate and functioning society requires its members to have a capacity to adopt others' perspectives. 
% 
% 
% 
%
% and achieve successful partnerships, 
%
% As Artificial Intelligence (AI) systems are 
%
% given increasingly sensitive and impactful roles in society, 
%
%
\commentsm{THIS IS A SHEILA COMMENT}\commentms{Without being too on the nose, I tried to emphasize that empathy is important and is a human skill and that AIs need it. I also tried to provide disclaimers as to what we do and do not do in this work.}

Critical to successful human interaction is a capacity for empathy - the ability to understand and share the thoughts and feelings of another.
As Artificial Intelligence (AI) systems are increasingly required to interact with humans in a myriad of settings, it is important to enable AI to wield empathy as a tool to benefit those it interacts with. In this paper, we work towards this goal by bringing together a number of important concepts: empathy, AI planning, and  reasoning in the presence of knowledge and belief\eat{epistemic reasoning}\eat{ and plan recognition (i.e., the problem of inferring an empathizee's plan and goal given observations about its behavior)}. We formalize the notion of \textit{Empathetic Planning}\eat{and \textit{Empathetic Plan Recognition}} which is informed by the beliefs and affective state of the empathizee. We appeal to an epistemic logic framework to represent the beliefs of the empathizee and propose AI planning-based computational approaches to compute empathetic solutions. We illustrate the potential benefits of our approach by conducting a study where we evaluate participants' perceptions of the agent's empathetic abilities and assistive capabilities.

\end{abstract}

\section{Introduction}\label{intro}

%\textbf{CHANGE EXAMPLE TO BUS GOING WRONG WAY EXAMPLE}\\

% “solved by using a different domain model and specific objectives about quality of solutions” - while our framework captures the preference-based planning setting you are describing, it also captures a much wider set of settings which cannot directly be modeled or solved using classical planning and plan recognition. 

% For example, consider that ACT is on a bus headed uptown but falsely believes the bus is headed downtown. In this scenario, an empathetic OBS should infer that ACT’s plan (taking a specific bus) will succeed under her (false) model of the world but will fail to achieve the goal (getting downtown) under some ground truth known to OBS. To solve the empathetic planning problem OBS must generate a multi-agent plan which involves conveying information about the bus to ACT (via an epistemic action that will change ACT’s beliefs about the world), followed by ACT getting off the bus and taking the correct one to downtown (prompted by the epistemic action). Our framework combines empathetic planning and PR and allows an empathetic agent to reason that ACT’s plan will achieve her goal in a way that is not optimal (or fail to achieve it altogether) and generate a better plan for ACT.\\

Artificial Intelligence (AI) systems are increasingly required to interact with other agents (be they human or artificial), but are still lagging in their ability to empathize with them when reasoning about their behavior. In the context of AI planning (the problem of selecting a goal-leading plan based on a high-level description of the world), an empathetic agent should be able to construct a plan that is harmonious with the goals, beliefs, values, affective state, and overall perspective of a fellow agent. Further, we are interested in facilitating the creation of empathetic agents holding a wide spectrum of roles - from passive observers, to virtual agents that offer relevant advice or act on someone's behalf, all the way to embodied agents who can physically interact with the environment and other agents in it. To illustrate, consider Alice who lives with panic disorder and agoraphobia. Alice \textit{fears} crowded places that might trigger a panic attack, and avoids busy restaurants, malls, and buses. Thus, Alice would never use public transit to get to work, despite it being the fastest way to get there. Instead, the optimal plan for her to get to work (that she can come up with on her own) addresses her fear, and would typically require her to walk instead of taking public transit, regardless of how suboptimal it might seem to an AI planning system bent on minimizing cost and time. A plan involving a crowded bus would simply not be executable by Alice. An empathetic AI that knows of an empty bus going along a similar route (which Alice \textit{believes} is always crowded or simply does not know about), could recommend to Alice that she take it instead of walking, which would help her save time and would not place her on a crowded bus.

% get to work in the downtown core, Alice would never take the crowded bus that is the fastest

% because she fears crowded buses

% Alice would never take the crowded bus  route to reach the downtown core because she fears crowded buses. 

% If you were to see a random occupant of Alice's building walking west, you would conjecture that they were heading to the park or a restaurant. If they walked east, you might conjecture they were heading towards public transit, and the fastest route to the downtown business core. Alice, however, would never take this route to reach the downtown core because she fears crowded buses. Instead, the optimal plan for her to get to work addresses her fear, and would typically require her to walk (in this case west) instead of taking public transit, regardless of how suboptimal it might seem to an AI planning system bent on minimizing cost and time. A plan involving a crowded bus would simply not be executable by Alice. The AI's ability to observe Alice's behavior and understand that it is, in fact, congruent with her fear is predicated on its empathy. An empathetic AI that knows of an empty bus going along a similar route (which Alice \textit{believes} is always busy), could recommend to Alice that she take it instead of walking.

To construct an empathetic AI that is able to empathize with Alice\eat{, understand her behavior,} and plan to assist her, we must provide it with a means of adopting her \textit{beliefs} and \textit{affective state}. To build towards this goal, our work brings together the notions of  empathy and epistemic planning, which is an emerging field of research that combines AI planning and reasoning in the presence of knowledge and belief.
\eat{
reasoning in the presence of knowledge and belief.\eat{. PR is the problem of inferring an agent's plan and goal given observations about its behavior}  \textit{Epistemic planning}}  We formalize the notion of \textit{Empathetic Planning} (EmP)\eat{and \textit{Empathetic Plan Recognition} (EmPR)} which builds on these concepts.
% %
% \eat{ As PR is critical to building assistive agents, our framework is not complete without a}
% %
% \eat{While the discussion in this work mostly focuses on EmP, we have also formalized the notion of \textit{Empathetic Plan Recognition} (EmPR) which is a core component of our integrative approach to planning and PR \commentms{refer to discussion in appendix?}.}
EmP requires an empathizer to empathize with an empathizee in order to construct plans that are faithful to the empathizee's view of the world. Specifically, we posit that in order to empathize with another, one must have at her avail a sufficiently rich representation and understanding of the beliefs and affective state of the agent with whom she is empathizing. Thus, some of the settings we address, involving reasoning about belief and affect, cannot directly be modelled as classical planning problems. We therefore appeal to a rich epistemic logic framework to represent the agent's beliefs and affective state. Lastly, we propose an epistemic planning-based computational approach to solving the EmP problem,
%
%**EmPR and EmP problems (and, importantly, their integration)**, 
%
thereby enabling the use of off-the-shelf epistemic planning tools. Our approach enables a sufficiently empathetic agent to generate a plan that is at least as `good' as the best plan the empathizee can generate by herself, using her own beliefs and capabilities. Finally, it is important to consider that a human's behavior does not always expose their intentions due to misconceptions or computational limitations. We submit that empathetic agents are well-suited for distinguishing between the underlying intent of the behavior and the actual performed behavior.
% \textbf{link to appendix which EMPR section and EMPR as epsitemic planning} while we formalize on both EmP and EmPR, we focus on EmP

% \textbf{say somewhere (PROBABLY IN THE PREVIOUS PARAGRAPH) that some of the settings we address require reasoning over the knowledge of another agent and cannot directly be modelled and solved using classical planning - maybe use } ``Our framework combines empathetic planning and PR and allows an empathetic agent to reason that ACT’s plan will achieve her goal in a way that is not optimal (or fail to achieve it altogether) and generate a better plan for ACT." 
% \\
% \textbf{OR USE THIS:} ``an empathetic agent can potentially will generate a plan that is at least as `good' as the best plan \textsc{Act} can generate by herself, using her own beliefs and capabilities."

Empathy is often thought of as \textit{the ability to understand and share the thoughts and feelings of another} and has an extremely rich history, beginning with its philosophical foundations and leading to research in fields such as psychology, ethics, and neuroscience (e.g., \cite{coplan2011empathy,davis2018empathy}). Empathy has been found to have two components, an affective, low-level component, and a cognitive, high-level component, with the two being interconnected \cite{shamay2011neural}. The affective component allows one to share the emotional experiences of another via affective reactions to their affective states. The cognitive component utilizes cognitive and affective Theory of Mind (ToM) - the ability to represent the mental states of others - and allows one to take the perspective of another, thereby facilitating reasoning over their \textit{mental} or \textit{affective} state. We focus on the cognitive component of empathy, and work towards building empathetic agents that can reason about the mental and affective states of other agents. We submit that pro-social AI agents should be equipped with a means of reasoning about the affective state of humans. This type of reasoning will lead to more socially acceptable behavior, as highlighted by recent work \cite{mcduff2018designing}. While we do not focus on affect in this work\eat{(and our brief treatment of it does not do justice with the rich existing body of work)}, our framework can be flexibly extended with various models of affect.
% we attempt to provide the reader with an intuition regarding future incorporation of affect into our framework.
%
Lastly, we note that while we aim to build \textit{assistive} empathetic agents that are benevolent, empathy can also facilitate malicious (or simply self-serving) motivations through manipulation. As such, the introduction of EmP encourages further exploration and discussion of these important areas.

There exists a large body of work on integrating empathy and ToM within intelligent agent systems in, e.g., psychological therapy, and intelligent tutoring \cite{mcquiggan2009modelling}. \cite{pynadath2005psychsim} created decision-theoretic ToM agents that can reason about the beliefs and affective states of other agents. However, this work has not appealed to the computational machinery of epistemic planning.\eat{ nor has it integrated planning and PR}\eat{ PR has also been researched extensively and was originally seen as an intersection of psychology and AI \cite{schmidt1978plan}.} Epistemic planning is an emerging field of research which is rapidly developing (e.g., \cite{baral2017epistemic}). For example, \cite{engesser2017cooperative} utilized epistemic planning and perspective taking to facilitate implicit coordination between agents. While the motivations driving their work and ours overlap, their work differs from ours both computationally and conceptually.
%does not propose to integrate planning and recognition.
%combine PR and planning, nor does it ** affect ; empathy? **
%
% While PR and planning have been combined in the past (e.g.,  \cite{talamadupula2014coordination,freedman2017integration}), previous work has not appealed to epistemic planning\eat{nor has it *****formalized empathy**** (or included affect)} in the context of such integration. 
Finally, the rich body of work on Belief-Desire-Intention (BDI) has studied affect and planning in the past (e.g., \cite{steunebrink2007logic}). However, BDI approaches have typically required agent plans to be specified in advance and we instead appeal to the flexibility of generative epistemic planning techniques to generate plans. Our approach is enabled by our combined knowledge of these fields of research and their decades-long development.

The main contributions of this paper are: (1) a formalization of EmP; (2) a computational realization of EmP that enables the use of existing epistemic planning tools; (3) a study which demonstrates the potential benefits of EmP in a diversity of domains.

\section{Preliminaries}
\commentms{Any way we can cut here without revs attacking us for doing so? it's already very densely packed}

In this section, we provide epistemic logic background and define the Multi-agent Epistemic Planning (MEP) problem\eat{ and describe some of the approaches to modelling it}.\eat{In the next section, we formally specify the notion of EPR, which will build upon the definitions in this section.}\eat{ We assume in the paper that agents act rationally (wrt their possibly false beliefs), i.e., they choose to perform actions which lead to an expected optimal result, where optimality is defined relative to some objective metric (in this work, the number of actions in a plan). Future work will explore various forms of limited reasoning.} We first present the multi-agent modal logic KD45$_{n}$ \cite{fagin2004reasoning} which we appeal to in our specification of EPR. Let $Ag$ and $\mathcal{P}$ be finite sets of agents and atoms, respectively.\eat{ We use $\phi$ and $\psi$ to represent formulae and $\top$ and $\bot$ to represent \textit{true} and \textit{false}, respectively.} The language $\mathcal{L}$ of multi-agent modal logic is generated by the following BNF:

\begin{equation} \nonumber
   \varphi ::= p \mid \lnot\phi \mid \phi \land \psi \mid B_i{\phi} 
\end{equation}

\noindent where $p \in \mathcal{P}$, $i \in Ag$, $\phi, \psi \in \mathcal{L}$ and $B_i{\phi}$ should be interpreted as \textit{``agent i believes $\phi$.''} We choose to represent the belief modality here so that we can model the false beliefs of agents. Using the equivalence $K_i{\phi} = B_i{\phi} \land \phi$, recent work on MEP has been able to capture both knowledge and beliefs.
%Further, recent work by \cite{huang2018general} and \cite{le2018efp} addressed common knowledge in the context of MEP and we choose not to focus on it here.
% For ease of exposition, we choose not to include the common knowledge modality, $C$, in the syntax. However, our specification of EPR can straightforwardly be extended to accommodate it.
\eat{A frame is defined as a pair $(W, R)$, where $W$ is a nonempty set of possible worlds. Further, for each agent $i \in Ag$, $R_i$ is a binary relation on $W$, called the accessibility relation for $i$.}
The semantics for formulae in $\mathcal{L}$ is given by Kripke models \cite{fagin2004reasoning} which are triplets, $M = \langle W, R, V \rangle$, containing a set of worlds, accessibility relations between the worlds for each of the agents ($R$ $=$ $\{R_i \mid i \in Ag\})$, and a valuation map, $V\colon W \to 2^{P}$\eat{, mapping worlds $w \in W$ to truth valuations}. When an agent $i$ is at world $w \in W$, $M$ determines, given the accessibility relations in $R_i$ pertaning to $w$, what worlds the agent considers possible.\eat{ A pointed Kripke model is a pair $s = (M, w)$, where $M$ is a Kripke model and $w$ is a world of $M$.} A formula $\phi$ is true in a world $w$ of a Kripke model $M = \langle W, R, V \rangle$, written $M,w \vDash \phi$, under these, inductively-defined conditions: $M, w \vDash p$ for an atom $p$ \textit{iff} $p$ $\in$ $V(w)$; $M, w \vDash \lnot\phi$ \textit{iff} $M, w \nvDash \phi$; $M, w \vDash \phi \land \psi$ \textit{iff} both $M, w \vDash \phi$; and $M, w \vDash \psi$, $M, w \vDash B_i{\phi}$ \textit{iff} $M, w' \vDash \phi \quad \forall w' \in W$ s.t. $R_{i}(w, w')$.
% \begin{itemize}
%     \item $M, w \vDash p$ for an atom $p$, \textit{iff} $p$ $\in$ $V(w)$,
%     \item $M, w \vDash \lnot\phi$, \textit{iff} $M, w \nvDash \phi$,
%     \item $M, w \vDash \phi \land \psi$, \textit{iff} both $M, w \vDash \phi$ and $M, w \vDash \psi$,
%     \item $M, w \vDash B_i{\phi}$, \textit{iff} $M, w' \vDash \phi \quad \forall w' \in W$ s.t. $R_{i}(w, w')$
% \end{itemize}
We say that $\phi$ is satisfiable if there is a Kripke model $M$ and a world $w$ of $M$ s.t. $M, w \vDash \phi$. Further, we say that $\phi$ entails $\psi$, written $\phi \vDash \psi$, if for any Kripke model $M$, $M, w \vDash \phi$ entails $M, w \vDash \psi$. Next, we assume some constraints on the Kripke model, with particular properties of belief, as discussed in \cite{fagin2004reasoning}. Namely, we assume that the Kripke model is \textit{serial} ($\forall w\,\exists v\,R(w\,,v)$), \textit{transitive} ($R(w\,,v)\wedge R(v\,,u)\Rightarrow R(w\,,u)$) and \textit{Euclidean} ($R(w\,,v)\wedge R(w\,,u)\Rightarrow R(v\,,u)$), with the resulting properties of belief: i. $B_i{\phi} \land B_i{(\phi \Rightarrow \psi)} \Rightarrow B_i{\psi}$ (K - Distribution); ii. $B_i{\phi} \Rightarrow \lnot B_i{\lnot \psi}$ (D - Consistency); iii. $B_i{\phi} \Rightarrow B_i{B_i{\psi}}$ (4 - Positive Introspection); and iv. $\lnot B_i{\phi} \Rightarrow B_i{\lnot B_i{\psi}}$ (5 - Negative Introspection). These axioms, together, form the KD45$_n$ system where $n$ signifies multiple agents in the environment. Note that the formal mechanisms for planning are described in the respective papers of the various off-the-shelf epistemic planners (e.g., \cite{huang2018general})\eat{ which are able to compute solutions to the EmP problem}. We provide a logical specification of the multi-agent epistemic planning problem that is not tied to a particular planner but rather provides a logical specification without embodying any of the syntactic restrictions that have been adopted by various planners.

\eat{\textbf{say this early one (yongmei mentions this in intro) and then reiterate after EPR def} 
The initial knowledge base (KB) and the goal, the preconditions and effects of actions can be arbitrary multi-agent epistemic formulas, progression of KBs wrt actions is achieved through higher-order belief revision or update based on effects of actions}

As mentioned, we appeal to epistemic planning, which combines AI planning and reasoning over the beliefs and knowledge of agents, to formally specify EmP. We appeal to a syntactic approach to epistemic planning (as opposed to a semantic one such as \cite{bolander2011epistemic}) and represent the initial Knowledge Base (KB) and other elements of the problem as arbitrary epistemic logic formulae. Further, we appeal to a multi-agent setting in order to represent the beliefs of the empathizer, the empathizee, and possibly other agents in the environment.

%\textbf{give preferences background here instead of in empathteic planning section - IF we're doing Preference-based Multi-agent Epistemic Planning (PMEP)}
\begin{definition}[MEP] \label{EPDef} A multi-agent epistemic planning problem is a tuple $\langle Q, I, G \rangle$ where $Q = \langle \mathcal{P}, \mathcal{A}, Ag \rangle$ is the MEP domain comprising sets of atoms $\mathcal{P}$, actions $\mathcal{A}$, and agents $Ag$, together with the problem instance description comprising the initial KB, $I \in \mathcal{L}$, and the goal condition $G \in \mathcal{L}$. 
 \end{definition}

To define how an action $a \in \mathcal{A}$ updates the state of the world in the epistemic planning framework, we follow \cite{huang2018general} who manipulates the KB with belief revision and update operators (e.g., \cite{alchourron1985logic}), $\circ$ and $\diamond$, respectively. A deterministic action is a pair $\langle Pre,\{(\gamma_1,\epsilon_1),...,
(\gamma_k,\epsilon_k)\}\rangle$, where $Pre \in \mathcal{L}$ is called the precondition of $a$, $\gamma_i \in \mathcal{L}$ is the condition of a conditional effect, and $\epsilon_i \in \mathcal{L}$ is called the effect of a conditional effect. A sensing action is a triplet $\langle Pre, pos, neg \rangle$, where $Pre, pos, neg \in \mathcal{L}$ are the precondition, the positive result, and the negative result, respectively. An action $a \in \mathcal{A}$ is executable wrt a KB $I \in \mathcal{L}$ if $I \vDash Pre(a)$. Suppose some deterministic action $a \in \mathcal{A}$ is executable wrt a formula $\phi \in \mathcal{L}$. \citeauthor{huang2018general} [\citeyear{huang2018general}] formally define the progression of $\phi$ wrt to a deterministic action $a \in \langle Pre, \{(\gamma_1,\epsilon_1),...,(\gamma_k,\epsilon_k)\}\rangle$ as follows: let \eat{$\{\gamma_i \mid \phi \vDash \gamma_i\}$} $\gamma_{i_{1}}, ..., \gamma_{i_{m}}$ be all the conditions of conditional effects s.t. $\phi \vDash \gamma_i$. Then $\phi'$, denoted by $prog(\phi,a)$, is a progression of $\phi$ wrt $a$ if $\phi' = ((\phi \diamond \epsilon_{i_{1}})...) \diamond \epsilon_{i_{m}}$, where $\epsilon_{i_{j}}$ is the effect corresponding to the condition $\gamma_{i_{j}}$.\eat{For example, let $p \in \mathcal{P}$, $i \in Ag$, $\mathcal{I} =  \lnot p \land B_i{\lnot p}$ and $a \in \mathcal{A}$ where $\epsilon_{j}$, the effect of one of $a$'s conditional effects, is $p \land B_i{p}$. The update operation, $\mathcal{I} \diamond \epsilon_{j}$, will yield $\mathcal{I}' = p \land B_i{p}$.} The progression of $\phi \in \mathcal{L}$ wrt a sensing action $a \in \mathcal{A}$ and positive result (resp. negative) is defined as $\phi^{+} = \phi \circ pos(a)$ (resp. $\phi^{-} = \phi \circ neg(a)$). Let $\phi \in \mathcal{L}$ and $\pi = a_1,...,a_k$ a sequence of actions. The progression of $\phi$ wrt $\pi$ (with sensing results for sensing actions) is inductively defined as follows: $prog(\phi,\epsilon)=\phi$; $prog(\phi,(a;\sigma)) = prog(prog(\phi,a),\sigma)$ if $\phi \vDash Pre(a)$, and undefined otherwise. $\epsilon$ is an empty sequence of actions. A \textbf{solution} to an MEP problem is an action tree branching on sensing results, such that the progression of the initial KB, $I$, wrt each branch in the tree entails the goal $G$.

% The formal mechanisms for planning (e.g., mutex, frame-axioms) are described in the respective papers of the various off-the-shelf epistemic planners (e.g., [Muise et al., 2015; Huang et al., 2018]) which we can use and which are able to generate plans that are solutions to the empathetic planning problem. Note that  
% %We will say that a sequence of actions achieves $G$ and solves the MEP problem interchangeably.

% Different solution concepts to MEP have been proposed, including: a linear plan \cite{le2018efp,MuiseBFMMPS15}; an action tree branching on the results of sensing actions \cite{huang2018general}; and an implicitly coordinated policy\eat{as the solution to the cooperative epistemic multi-agent planning problem} \cite{engesser2017cooperative}. We focus on linear plans and define a \textbf{solution} to an MEP problem as a sequence of actions (a plan), $\pi = a_1,...,a_k$, such that $prog(\mathcal{I}, \pi) \vDash G$. 
%Finally, the \textit{cost} of $\pi$ is the number of actions it contains.
\section{Empathy}

In this section, we discuss the notion of empathy, contrast it with sympathy, and discuss possible characteristics of \textit{assistive empathetic agents}.
As mentioned in Section \ref{intro}, inspired by the rich history of empathy and its philosophical foundations, we define empathy as \textit{the ability to understand and share the thoughts and feelings of another}. When taken to the extreme, where an empathizer knows every thought and feeling of an empathizee, the above definition embodies an idealized notion of empathy which necessitates \textit{omniscience}. We later discuss more pragmatic notions of empathy, which only require an empathizer to be empathetic `enough' and are defined wrt a specific task. To contrast, consider the Golden Rule, which asks us to treat others as \textit{we} would like to be treated. The rule assumes similarity, implying that others would like to be treated the same way we would and, therefore, does not allow for the existence of multiple perspectives, thus leading to \textit{sympathetic} behavior \cite{bennett1979overcoming}. Empathy, on the other hand, allows an \textit{empathetic agent} to experience the world from the perspective of the empathizee. In Section \ref{eval}, we conduct a study and compare an empathetic agent and a sympathetic agent.

% \textbf{As an example, consider... therapist?}

\eat{ This is because the belief consequences for their KBs will be equivalent.} \eat{Interestingly, we say that \textsc{Act} is \textit{self-understanding} $\iff$ \textsc{Act} empathizes with \textsc{Act} wrt to $I^{\textsc{Act}}$ and $I_{true}^{\textsc{Act}}$ (\textsc{Act}'s so-called ground truth KB, that may not be accessible to \textsc{Act}).}
% Next, we note that $I^{\textsc{Act}}$ is \textsc{Act}'s projection of its own beliefs, while $I_{true}^{\textsc{Act}}$ is the agent's \textsc{Act}'s so-called ground truth KB, that may not be accessible to \textsc{Act}\eat{ (let alone \textsc{Obs})}. We say that 
%  We say that \text{Obs} \textit{sympathizes} with \text{Act} \textit{iff} for any $\phi \in \mathcal{L}$, $I \vDash \phi$ iff $I^{proj} \vDash \phi$ and for any $\psi \in \mathcal{L}$ and any executable action sequence $\sigma = a_1,...,a_k$, $prog(I,\sigma) \vDash \psi$ iff $prog(I^{proj}, \sigma) \vDash \psi$. That is, \textsc{Obs} 

% \textbf{Explain that symp agent has some normative model and that's why it's bad?}

\subsubsection*{Affect}
As mentioned, cognitive empathy includes both cognitive and affective components. If Emily empathizes with Alice, she should be able to reason about Alice's affective state. There is an extremely rich body of work on theories of affect (e.g., \cite{lazarus1966psychological}) and on the incorporation of these theories into computational models of affect (e.g., \cite{gratch2000emile}) Further, previous work has formalized logics of emotion for intelligent agents. E.g., \citeauthor{steunebrink2007logic} [\citeyear{steunebrink2007logic}] defined a formal logic within a BDI framework while \citeauthor{lorini2011logic} [\citeyear{lorini2011logic}] use a fragment of \textsc{STIT} logic to formalize counterfactual emotions such as regret and disappointment. Both of these logics appeal to a notion of epistemics to formalize complex emotions which are predicated on the beliefs of agents. Previous work has also quantified emotional intensity using fuzzy logic \cite{el2000flame}. Thus, there exists a multitude of ways in which to integrate affect into our framework, and as part of future work we will experiment with various extensions of our approach. 

% and in the future we will experiment with various extensions of our approach

% While we do not focus on affect in this work, we wish to provide the reader with an intuition regarding the incorporation of affect into an empathetic planning and PR framework.

% \eat{, including different affective components (e.g., personality and mood) and the interplay between them}. 

% To give an intuition for how affect can be incorporated into our framework, we focus on a widely-accepted correspondence between affect and behavior. Specifically, we formalize the notion of fear (as mediated by agoraphobia and a panic disorder) as a cause of avoidant behavior \cite{american2013diagnostic}. Note that in our realization of EmPR and EmP, we assume that the affect theory in use is compiled into the action theory, and thus reflected in $\mathcal{A}$, the set of actions. 

% E.g., one of the effects of the action \textit{boardFullBus(agent, bus)} is \textit{inCrowd(agent)}. Since the source of Alice's fear is being in a crowd, she will avoid full buses. Therefore, in our simplified realization, \textit{boardFullBus(agent, bus)} is removed from $\mathcal{A}$ and excluded from all generated plans.

\subsubsection*{Assistive Empathetic Agents}
Before formalizing the notion of EmP, we discuss some of the properties\eat{assumptions/postulates//desiderata} we believe should characterize an assistive empathetic agent. More accurately, these are agents who wish to empathize with another, as empathy is often thought of as an ongoing process (or mountain climb \cite{ickes1997empathic}). \textbf{i. Need for Uncertainty\quad}\eat{The computational approaches described in Section \ref{comp} assume that \textsc{Obs} empathizes with \textsc{Act}. However, when this is not the case, the quality of inferences may suffer. }\eat{ \citeauthor{hadfield2017off} [\citeyear{hadfield2017off}] propose a scenario where a robot seeks to maximize and align with human values, but is also uncertain about them. Thus, the robot has an incentive to keep its off-switch enabled as it reasons that it must be doing something `wrong' if it is being turned off. Similarly, } We posit that an agent who wishes to empathize with another should be uncertain about the empathizee's values (e.g., \cite{hadfield2017off}) and, importantly, about her true beliefs about the world. Since the quality of plans generated by an assistive empathetic agent is predicated on the veracity of the empathetic agent's model of the empathizee, the empathetic agent should unceasingly strive to align itself better with her. \textbf{ii. Benevolence\quad} An assistive empathetic agent should take it upon itself to benefit the empathizee in a way that is aligned with the latter's values. \textbf{iii. $\lnot\text{Machiavellianism}$\quad} As mentioned, empathy can facilitate malicious intent. Thus, an assistive empathetic agent should not be Machiavellian, i.e., use deception, manipulation, and exploitation to benefit its interests. Interestingly, empathy and Machiavellianism have been found to be negatively correlated \cite{barnett1985role} - i.e., the will to manipulate is present in Machiavellians, but the means by which to do so are often not. Relatedly, a study conducted by \citeauthor{botsLie2019} [\citeyear{botsLie2019}] showed that human participants were, in general, positive towards an AI agent lying, if it was done for the `greater good'. Such questions should be explored further. E.g., when interacting with humans who wish themselves (or others) harm.

\section{Empathetic Planning}
\label{emp_plan}

In this section, we define the EmP problem and its solution. We will make use of the notation \textsc{Obs} (observer) for the empathizer  and \textsc{Act} (actor) for the empathizee throughout the paper; both are assumed to be in the set of agents $Ag$ in all definitions.
To be helpful, an assistive empathetic agent should be able to reason about the \textit{preferences} of the empathizee. In general, agent preferences can be specified in various ways, including a reward function over states, as is done in a Markov Decision Process (MDP), or by encoding preferences as part of a planning problem (e.g., \cite{jorge2008planning}). Preferences (or rewards) can also be augmented with an emotional component which can encode an aversion to negative emotions, while optimizing for positive ones \cite{moerland2018emotion}. To simplify the exposition, we focus on plan costs as a proxy for agent preferences. We assume in this work that we have an accurate approximation of \textsc{Act}'s preferences over plans (in this work, the lower the cost, the better the plan), and do not focus on the problem of acquiring a faithful approximation of an agent's preferences.

 \begin{definition}[EmP] An empathetic planning problem is a tuple $Z$ $=$ $\langle Q, I, {G}_{\textsc{Act}} \rangle$, where $Q$ $=$ $\langle \mathcal{P}, \mathcal{A}, Ag \rangle$ is an MEP domain, $\mathcal{P}$, $\mathcal{A}$, and $Ag$ are sets of atoms, actions, and agents, respectively, $I \in \mathcal{L}$ is the initial KB, and ${G}_{\textsc{Act}} \in \mathcal{L}$ is \textsc{Act}'s estimated goal.
\label{empathetic_planning}
 \end{definition}
%   \begin{definition}[EmP] An empathetic planning problem is a tuple $Z$ $=$ $\langle Q, I, {G}_{\textsc{Act}}\rangle$, where $Q$ $=$ $\langle \mathcal{P}, \mathcal{A}, Ag \rangle$ is an MEP domain, $\mathcal{P}$, $\mathcal{A}$, and $Ag$ are sets of atoms, actions, and agents, respectively, $I \in \mathcal{L}$ is the initial KB, ${G}_{\textsc{Act}} \in \mathcal{L}$ is \textsc{Act}'s estimated goal, and $\preceq$ is a transitive and reflexive relation in $\Pi$, where $\Pi$ contains precisely all solutions to $\langle Q, I, {G}_{\textsc{Act}} \rangle$.
% \label{empathetic_planning}
%  \end{definition}
%  \textbf{perhaps use the ALICE - model the problem as in EPR project and show solution that ACT can come up with and better solution that OBS can come up with}

 To illustrate, we partially model the example from Section \ref{intro} as an EmP problem (with Alice as \textsc{Act}).

\begin{itemize}
    \item $I = $  $B_{\textsc{Obs}}$ \textit{at(\textsc{Act}, home)} \\$\land$  $B_{\textsc{Obs}}$ \textit{travelsBetween(alternativeBus, home, work)} \\ $\land$
    $B_{\textsc{Obs}}$ \textit{crowded(alternativeBus)} \\ $\land$
    $B_{\textsc{Obs}}$ $B_{\textsc{Act}} \lnot$ \textit{crowded(alternativeBus)} $\land$ ...
    \item ${G}_{\textsc{Act}} =$ \textit{at(\textsc{Act}, work)}
\end{itemize}

% An action $a \in \mathcal{A}$ is defined as in the previous section. Next, we define a solution to the EmP problem:

 % While in this line of work, we are concerned more with the generation of the content of explanations rather than the actual delivery of this information, there has been some recent work to this end. Depending on the type of interaction between the planner and the human, this can be achieved by means of natural language dialog (Perera et al. 2016), in the form of a graphical user interface (Sengupta et al. 2017; Chakraborti et al. 2018b) or even in mixed-reality interfaces (Chakraborti et al. 2018d; 2018c). \textbf{FROM human-aware planning revisited

 \begin{definition} Given an EmP problem, $Z$ $=$ $\langle Q, I, {G}_{\textsc{Act}}\rangle$, $\pi$ is an \textbf{assistive solution} to $Z$ \textit{iff} $\pi$ solves $\langle Q, I, {G}_{\textsc{Act}} \rangle$ and $\pi \in \Pi^{*}$ where $\Pi^{*}$ is the set of optimal solutions for $\langle Q, I, {G}_{\textsc{Act}}\rangle$.
 %
%  \{ \pi' \in \Pi \mid \text{there is no}\enskip \pi'' \in \Pi \enskip \text{such that}\enskip \pi'' \prec \pi' \}$.
 %
\label{empathetic_planning_solution}
 \end{definition}
 
%   \begin{definition} Given an EmP problem, $Z$ $=$ $\langle Q, I, {G}_{\textsc{Act}}\rangle$, a plan $\pi$ that solves $\langle Q, I, {G}_{\textsc{Act}} \rangle$ is an assistive \textbf{solution} to $Z$ \textit{iff} $\pi \in \{ \pi' \in \Pi \mid \text{there is no}\enskip \pi'' \in \Pi \enskip \text{such that}\enskip \pi'' \prec \pi' \}$.
% \label{empathetic_planning_solution}
%  \end{definition}

The solution to the EmP problem may vary if \textsc{Obs} is not trying to help \textsc{Act} achieve her goal (e.g., adversarial interaction as in \cite{freedman2017integration}). Throughout the paper, we focus on finding assistive solutions.

We now formally define a pragmatic notion of empathy, defined wrt the EmP problem. To this end, we first discuss the notion of \textit{projection}, where an agent can reason from another agent's perspective. Let $\langle \langle \mathcal{P}, \mathcal{A}, Ag \rangle, I \rangle$ be \textsc{Obs}'s MEP domain and initial KB. $\mathcal{A}$ and $I$ encode \textsc{Obs}'s beliefs about the world, including, importantly, its beliefs about \textsc{Act}'s beliefs about the world. To enable \textsc{Obs} to reason from \textsc{Act}'s perspective, we wish to project $\mathcal{A}$ and $I$ wrt to $\textsc{Act}$. We define the \textbf{projection} of a formula $\phi$ with respect to an agent $i$, $proj(\phi,i)$. Given $\phi,\psi \in \mathcal{L}$, and assuming $\phi$ and $\psi$ are in NNF form, $proj(\phi,i)$ $=$ $\psi$ when $\phi = B_{i}\psi$ and is undefined otherwise. Both \cite{MuiseBFMMPS15} and \cite{engesser2017cooperative} similarly define projection operators, with \citeauthor{MuiseBFMMPS15}'s syntactic approach being similar to ours, while \citeauthor{engesser2017cooperative} define a semantic equivalent. We project $\mathcal{A}$ and the closure of $I$ (defined as $\{\psi \in \mathcal{L} \mid I \vDash \psi\}$) wrt \textsc{Act}. Note that the closure will be infinite but for any practical computation we will limit generation of closure to the relevant subset. We project every formula in the closure of $I$ and for every $a \in \mathcal{A}$ we project every precondition, and for every conditional effect of $a$ we project the condition and effect (we project the positive and negative results for sensing actions). We refer to the result of the projection operation wrt $\mathcal{A}$ and $I$ as $\mathcal{A}^{proj}$ and $I^{proj}$, respectively.

Let $Z$ $=$ $\langle Q, I, {G}_{\textsc{Act}}\rangle$ be an EmP problem. Let $\Pi_{proj}^{*}$ and $\Pi_{\textsc{Act}}^{*}$ be the sets of optimal solutions among $\Pi_{proj}$ and $\Pi_{\textsc{Act}}$, respectively, where $\Pi_{proj}$ and $\Pi_{\textsc{Act}}$ are the sets of all solutions for $\langle \langle \mathcal{P}, \mathcal{A}^{proj}, Ag \rangle, I^{proj}, {G}_{\textsc{Act}}\rangle$ and $\langle \langle \mathcal{P}, \mathcal{A}^{\textsc{Act}}, Ag \rangle, I^{\textsc{Act}}, {G}_{\textsc{Act}}\rangle$, respectively. $\langle \mathcal{P}, \mathcal{A}^{\textsc{Act}}, Ag \rangle$ and $I^{\textsc{Act}}$ are \textsc{Act}'s true MEP domain and initial KB, which are typically not accessible to \textsc{Obs}.\eat{ We use $Q^{proj}$ as shorthand for $\langle \mathcal{P}, \mathcal{A}^{proj}, Ag \rangle$.} We say that $\textsc{Obs}$ is \textbf{\textit{selectively task-empathetic}} wrt to \textsc{Act} and $Z$ \textit{iff} $\Pi_{proj}^{*}$ $=$ $\Pi_{\textsc{Act}}^{*}$. That is, \textsc{Obs} needs to be empathetic `enough' in order to generate, when projecting to reason as \textsc{Act}, the optimal solutions that achieve ${G}_{\textsc{Act}}$ and which \textsc{Act} can generate on her own (using only her beliefs and capabilities). Importantly, if \textsc{Obs} is selectively task-empathetic wrt $Z$, she will generate a plan that is at least as `good' as the best solution \textsc{Act} can generate by herself, using her own beliefs and capabilities. 

\commentms{Possibly say: We prove this result in the appendix.}

% \begin{pro}
% Given an EmP problem $Z$ $=$ $\langle Q, I, {G}_{\textsc{Act}}, \preceq\rangle$, \textsc{Obs} is selectively task-empathetic wrt to \textsc{Act} and $Z$ if \textsc{Obs} empathizes with \textsc{Act}.
% \end{pro}
%
\eat{Since we are building assistive empathetic agents, $\pi$ should be \textit{executable} by \textsc{Act} (perhaps jointly with \textsc{Obs}).}
\textsc{Obs}' solution could be better than \textsc{Act}'s solution if \textsc{Obs} has, for instance, additional knowledge or capabilities that \textsc{Act} lacks. In this case, a solution to the EmP problem, $\pi$, may solve $\langle \langle \mathcal{P}, \mathcal{A}, Ag \rangle, I, {G}_{\textsc{Act}}\rangle$ but not $\langle \langle \mathcal{P}, \mathcal{A}^{proj}, Ag \rangle, I^{proj}, {G}_{\textsc{Act}}\rangle$.
%
% executable in $I$ but not in $I^{proj}$\eat{, and also be better than any plan that \textsc{Act} can generate on her own }. 
%
% executable in $\langle Q^{proj}, I^{proj} \rangle$\eat{, the projection of $\langle Q, I \rangle$ wrt \textsc{Act}}. However, $\pi$ could also be a plan that is 
%
Returning to our example, recall that \textsc{Act} avoids crowded buses and so will not board the crowded bus that goes from her home to work. Thus, the best plan \textsc{Act} can come up with is walking from home to work (assuming there is no alternative mode of transportation) since she believes the alternative bus is crowded. However, since \textsc{Obs} knows of an alternative bus that is relatively empty\eat{(assuming \textsc{Obs} is correct in its beliefs)}, a better plan would include \textsc{Act} taking the alternative bus. However, to make this plan executable in \textsc{Act}'s model, \textsc{Obs} must inform \textsc{Act} that the bus is empty (e.g., \textit{inform(\textsc{Obs}, \textsc{Act}, busInfo)}). In this case, \textsc{Act}'s goal of getting to work, while ontic, requires epistemic actions such as informing \textsc{Act} of the bus' status and the underlying MEP framework can facilitate reasoning about the required epistemic action(s) to achieve the goal. To contrast, consider Sympathetic \textsc{Obs}, who assumes that \textsc{Act} shares its model of the world. Thus, the optimal plan that solves $\langle Q, I, {G}_{\textsc{Act}} \rangle$ consists of \textsc{Act} taking the crowded bus, which is a plan that is not executable in her true model, due to her panic disorder.

\section{Computation} \label{comp}

In this section, we describe how to compute a solution to the EmP problem\eat{, as well as how to integrate the two paradigms}. In our computation and evaluation, we assume that \textsc{Obs} is omniscient\eat{(and thus also empathizes with \textsc{Act})}. In this case, a solution to an MEP problem\eat{, an action tree,} `collapses' to a single path, as \textsc{Obs} has knowledge of the results of the sensing actions. We call the sequence of actions that defines the path a plan and say that its cost is the number of actions in the sequence.

\subsubsection*{Computing a Solution to the EmP Problem}
A solution to the EmP problem, $Z$ $=$ $\langle Q, I, {G}_{\textsc{Act}}\rangle$, is a plan $\pi$ that belongs to the set of optimal plans which achieve ${G}_{\textsc{Act}}$, $\Pi^{*}$.\eat{To realize this solution computationally, we turn, once again, to epistemic planning.} While ${G}_{\textsc{Act}}$ can be obtained in different ways (e.g., \textsc{Act} could explicitly tell \textsc{Obs} their goal or communicate it to a fellow agent), inferring ${G}_{\textsc{Act}}$ via plan recognition (i.e., the problem of inferring an actor's plan and goal given observations about its behavior) is of most interest to us.
%
%
% . Since we are interested in the integration of planning and recognition, we can infer ${G}_{\textsc{Act}}$ using PR.
%
% As mentioned, we do not focus on EmPR in this work. \commentms{something like this?}
%
% and since we are interested in the integration of EmPR with EmP, we *** focus *** on inferring ${G}_{\textsc{Act}}$ using plan recognition. **As mentioned, we do not focus on EmPR in this work and more details can be found in the Appendix we have linked to.*** %
At an overview, by solving the recognition problem we can set $G_{\textsc{Act}}$ to be the goal most likely being pursued by \textsc{Act}, given a sequence of observations. While we do not focus on the recognition component in this work, we have formalized the notion of empathetic plan recognition in an unpublished manuscript and proposed an integrative approach to empathetic planning and plan recognition. Further investigation is left to future work.
% More details on the formalization of EmPR and its integration with EmP can be found in the appendix.\commentms{something like this?}
% Given an EmPR problem, $R$ $=$ $\langle Q, I, \mathcal{G}, O \rangle$, we use the EmPR as Planning approach and compute $P(G|O)$. ${G}_{\textsc{Act}}$ is then the goal deemed most likely wrt $P(G|O)$ (the full distribution over $\mathcal{G}$ can be addressed by introducing necessities of propositions \cite{freedman2017integration}). 
%
To obtain a solution to the EmP problem, we solve the MEP problem $\langle Q, I, {G}_{\textsc{Act}}\rangle$ by using an optimal off-the-shelf epistemic planner.\eat{There are currently no off-the-shelf epistemic planners that readily address preferences, and so in our experiments $\preceq$ is defined wrt the cost of a plan, where $\pi_1$ is preferred to $\pi_2$ \textit{iff} $\pi_1$ has a lower cost than $\pi_2$.}\eat{ As described in the next section, we use an off-the-shelf optimal epistemic planner to solve $\langle Q, I, {G}_{\textsc{Act}}\rangle$.}

\textbf{Complexity\quad}\commentms{I wonder if we can do without this paragraph. 2 of the reviewers mentioned complexity}Epistemic planning has been shown to be computationally expensive (e.g., \cite{aucher2013undecidability}).\eat{ There are, however, a number of decidable and expressive fragments of epistemic planning (e.g., \cite{charrier2016impact}).} E.g., the encoding process in RP-MEP, the epistemic planner proposed by \cite{MuiseBFMMPS15} and used in this work, generates an exponential number of fluents when transforming the problem into a classical planning problem.

\section{Evaluation}\label{eval}

In our preliminary evaluation, we set out to (1) expose the diversity of tasks that can be captured by EmP (2) demonstrate that existing epistemic planners can straightforwardly be used to solve EmP problems and (3) to evaluate the benefits of our approach as assessed by humans. To this end, we constructed and encoded a diversity of domains, ran them using an off-the-shelf epistemic planner and conducted a study with human participants. For all of our experiments, we used the latest version of the epistemic planner RP-MEP \cite{MuiseBFMMPS15} with the Fast Downward planner \cite{helmert2006fast} with an admissible heuristic. Note that various epistemic planners impose various restrictions on domain modeling and plan generation, and we will experiment with different planners in the future.\eat{ We first describe some of the scenarios .} The various scenarios used in our simulations and study represent a diversity of every-day situations which illustrate the potential benefits of our approach. Further, the scenarios involve Alice (as \textsc{Act}) and her virtual assistant (as \textsc{Obs}), including suggestions (automatically generated solutions to an EmP problem) given to Alice by \textsc{Obs}. We then present the results of the study. We encode all scenarios as EmP problems, including the beliefs of \textsc{Obs} and \textsc{Act} and \textsc{Act}'s goal, $G_{\textsc{Act}}$ (see example in Section \ref{emp_plan}). We run RP-MEP once to compute a solution to an EmP problem.
\commentms{Refer to appendix for EmPR results?}\\
\commentms{Refer to appendix for SurveyMonkey questionnaire?}

% , which is then presented to the particiapant

% \textbf{Simulating the Scenarios\quad} 

% In some of the scenarios, $G_{\textsc{Act}}$ is the inferred goal (the most likely goal wrt $P(G|O)$) of an EmPR problem which includes a set of possible goals, and a sequence of observations that we sample from an optimal hidden plan. To compute $\Delta$, we transform the EmPR problem as described in Section \ref{comp} and run RP-MEP twice on the transformed problem - with and without satisfying $O$.  \textbf{emphasize that we automatically produce solutions to EmP problem using generative epistemic planning tools and that the generated solution in then presented to the participants.}

\textbf{Experimental Setup\quad} The study aims to evaluate both perceptions of the agent's empathetic abilities and perceptions of the agent's assistive capabilities, as assessed by humans. To test this, participants were presented with 12 planner-generated textual scenarios (some of which are presented below) of either empathetic or sympathetic agents and were asked to rate the following two claims pertaining to each scenario on a 5 point scale ranging from strongly disagree to strongly agree: (1) \textit{``The virtual assistant was able to successfully take Alice's perspective"} (reflecting our measure of participants' perceptions of the agent's empathetic abilities); and (2) \textit{``If I were Alice, I would find this virtual assistant helpful"} (reflecting our measure of participants' perceptions of the agent's assistive capabilities). Scenarios with empathetic agents and scenarios with sympathetic agents were identical, except for the EmP solution generated by the agent. The sympathetic agent assumes that Alice shares its model of the world when computing a solution to the EmP problem. We had a total of 40 individuals (28 female) participate, ranging from 18 to 65 years old. Participants were recruited and completed the questionnaire via an online platform, and had no prior knowledge about the study.
%, and were compensated. 

\textbf{Scenario 1\quad} Alice is on a bus headed uptown but believes the bus is headed downtown. \textsc{Obs} suggests that Alice get off the bus and get on the correct one. As claimed in Section \ref{intro}, empathetic agents are well-suited for distinguishing between the underlying intent of the behavior and the actual performed behavior. In this scenario, \textsc{Obs} can infer that Alice's current plan (riding the wrong bus) will not achieve her goal of getting downtown, the underlying intent of her behavior\eat{ (her goal could have been inferred from, for instance, a text message she sent to a friend she is supposed to meet downtown)}.
%
%\textsc{Obs} could also infer, perhaps based on 
%
% We submit that empathetic agents are well-suited for distinguishing between the underlying intent of the behavior and the actual performed behavior.
%
% Alice's current plan (staying on her current bus) will not achieve her goal. This plan, this, has infinite cost. Therefore, a plan involving telling Alice about the wrong bus and making her get off and take another bus will have lower cost.
%
\textbf{Scenario 2\quad} Alice is visiting her 91-year-old grandmother, Rose. Rose cannot hear very well but feels shame when asking people to speak up. \textsc{Obs} detects that Alice is speaking softly and sends her a discreet message suggesting she speak louder. Alice's goal is for her grandmother to hear her. This goal could be achieved by Rose asking Alice to speak louder. However, this plan will have a higher cost (due to Rose's aversion to asking people to speak louder) than the plan involving \textsc{Obs} sending a discreet message to Alice. \textbf{Scenario 3\quad} This scenario models the example from Section \ref{intro}, where Alice is trying to get to work and avoids crowded buses. \textsc{Obs} knows about a relatively empty bus, which would save her time, and suggests that Alice take it to work (in the sympathetic case, \textsc{Obs} suggests the crowded bus).
\textbf{Scenario 4\quad} Alice is trying to get to her friend's house and there are two ways leading there - one is well-lit and populated (but slower) while the other is dark and deserted. Alice prefers to feel safe when walking outside after dark. \textsc{Obs} suggests Alice take the well-lit route (compared to the dark route in the sympathetic case). \eat{Neither Rose nor Alice want to feel shame (other person) }

% ***Study-1 aims to develop an understanding of how humans respond to the task of generating explanations, i.e. if left to themselves, humans preferred to generate explanations similar to the ones enumerated in Section 3.2. To test this, we asked participants to assume the role of the internal agent in the explanation process and explain their plans with respect to the faulty map of their teammate.*** The experimental setup included 9 scenarios where the participants were asked to rate two claims pertaining to the scenario: (1) \textit{``The virtual assistant was able to successfully take Alice's perspective"}; and (2) \textit{``If I were Alice, I would find this virtual assistant helpful"}. The answers to these questions were measured using a five-point Likert scale (**ranging from Strongly Disagree to Strongly Agree**). The rating of the first claim measures .... and the rating of the seconds question measures .... In total, we had 40 participants, 12 male and 28 female, between the age range of 18-65. Participants for the study were recruited via an online platform, had no prior knowledge about the study, and were paid to participate.

\textbf{Results\quad} 
Across the 9 scenarios containing empathetic agents, 87\% of participants either strongly agreed or agreed with statement (1) and 82\% either strongly agreed or agreed with statement (2). Results comparing participants' perceptions of the agent's empathetic abilities demonstrated a statistically significant difference in ratings of scenarios containing empathetic agents (\textit{M}=4.21, \textit{SD}=1.01), compared to sympathetic agents (\textit{M}=1.72, \textit{SD}=0.73), \textit{t}(39)= 12.48, \textit{p} $<$ 0.01.

\section{Discussion and Summary}

In this work, we have introduced the notion of EmP which we formally specified by appealing to a rich epistemic logic framework and building upon epistemic planning paradigms. We proposed a computational realization of EmP as epistemic planning that enables the use of existing epistemic planners. We conducted a study which demonstrated the potential benefits of assistive empathetic agents in a diversity of scenarios as well as participants' favorable perceptions of the empathetic agent's assistive capabilities. It has been claimed recently that richer representational frameworks are needed to bridge some of the gaps in current virtual assistants - AI systems that frequently interact with human users \cite{cohen2018back}. The scenarios we have presented illustrate precisely this need in the context of reasoning over the beliefs and goals of agents and our approach is well-suited to address such settings. 

% , as well as our approach's ability to

% Our work demonstrates exactly that by introducing empathetic and assistive reasoning into **these systems. **

% It is evident that existing virtual assistants, , have a far way to go to become virtual companions who proactively assist the user by using the latter's mental and affective state. 

\eat{Further, while we focus on assistive empathetic agents, empathy may be wielded by agents who are not benevolent and this `darker side' of empathy invites a deeper discussion of EmP and EmPR.}

There is a diverse body of research related to the ideas presented here. Empathy has been incorporated into intelligent agent systems in various settings (e.g., \cite{aylett2005fearnot}) but has not appealed to generative epistemic planning techniques. Work on BDI has explored notions of epistemic reasoning \cite{sindlar2008mental} and affect \cite{steunebrink2007logic}, but has typically not appealed to the flexibility of AI planning to generate plans.\eat{\citeauthor{freedman2017integration} [\citeyear{freedman2017integration}] have integrated planning with PR but haven't addressed agent beliefs or affect. Finally, \cite{talamadupula2014coordination} have combined belief modelling with PR but did not address affect or epistemic goals.} There exist many avenues for future work such as the integration of affect into our framework. Future work could also appeal to related work which attempts to reconcile the human's model \cite{chakraborti2017plan} when providing the human with explanations (e.g., informing them of the bus' status). Lastly, partially observable settings will be addressed, where the empathtizer may have uncertainty regarding the mental state of the empathizee, as well as the environment in general.

\subsubsection*{Acknowledgments} 

The authors gratefully acknowledge funding from NSERC and thank their colleague Toryn Q. Klassen for helpful discussions.

%\pagebreak
\eat{
\subsection*{Acknowledgements}
We thank Karin Niemantsverdriet for her much appreciated counsel and support with the visual elements of this paper.
This material is based upon work supported in whole or in part with funding from 
the Laboratory for Analytic Sciences (LAS). Any opinions, findings, conclusions, 
or recommendations expressed in this material are those of the author(s) and do 
not necessarily reflect the views of the LAS and/or any agency or entity of the 
United States Government. The third author gratefully acknowledges funding from the Natural Sciences and Engineering Research Council of Canada (NSERC).
}
%Potentially shorter Acknolwedgemetns
%~~

\eat{
\noindent
{\bf Acknowledgements: } We gratefully acknowledge Karin Niemantsverdriet's assistance with the design and creation of figures; and the Laboratory for Analytic Sciences (LAS) and the Natural Sciences end Engineering Research Council of Canada (NSERC) for funding this work. Any opinions, findings, conclusions, 
or recommendations expressed in this material are those of the author(s) and do 
not necessarily reflect the views of NSERC, the LAS and/or any agency or entity of the 
United States Government. 
% Can this be shortened to "US Gov't" ?
}

%\pagebreak

\bibliographystyle{named}

\begin{small}
{\fs{9}
%{\small
\bibliography{bib}
}
\end{small}

\end{document}